\documentclass{article}

\usepackage{microtype}
\usepackage{graphicx}
\usepackage{amsfonts}
\usepackage{amsmath}
\usepackage{amssymb}
\usepackage{subfigure}
\usepackage{booktabs} % for professional tables
\usepackage{multirow}

\usepackage{hyperref}

\usepackage[accepted]{icml2021}

\def\eqref#1{equation~\ref{#1}}
\def\1{\bm{1}}

\DeclareMathAlphabet{\mathsfit}{\encodingdefault}{\sfdefault}{m}{sl}
\SetMathAlphabet{\mathsfit}{bold}{\encodingdefault}{\sfdefault}{bx}{n}

\newcommand{\Ep}[2]{\mathbb{E}_{#1}\left[\,#2\,\right]}

\newcommand{\ex}{\mathbb{E}}

\icmltitlerunning{}

\begin{document}

\twocolumn[
\icmltitle{Adapting Double Q-Learning for Continuous Reinforcement Learning}

\icmlsetsymbol{equal}{*}

\begin{icmlauthorlist}
\icmlauthor{Arsenii Kuznetsov}{independent}
\end{icmlauthorlist}

\icmlaffiliation{independent}{Independent researcher, Tbilisi}

\icmlcorrespondingauthor{Arsenii Kuznetsov}{brickerino@gmail.com}

\icmlkeywords{Machine Learning}

\vskip 0.3in
]



\printAffiliationsAndNotice{}  % leave blank if no need to mention equal contribution

\begin{abstract}
Majority of off-policy reinforcement learning algorithms use overestimation bias control techniques.
Most of these techniques rooted in heuristics, primarily addressing the consequences of overestimation rather than its fundamental origins.
In this work we present a novel approach to the bias correction, similar in spirit to Double Q-Learning.
We propose using a policy in form of a mixture with two components.
Each policy component is maximized and assessed by separate networks, which removes any basis for the overestimation bias.
Our approach shows promising near-SOTA results on a small set of MuJoCo environments.
% In this work, we present a data-driven approach for automatic bias control.
% We demonstrate its effectiveness on three algorithms: Truncated Quantile Critics, Weighted Delayed DDPG and Maxmin Q-learning. 
% Our approach eliminates the need for an extensive hyperparameter search.
% We show that it leads to the significant reduction of the actual number of interactions while, in most cases, matching the performance of a  resource demanding grid search method.
% While on average the reduction of the bias improves the performance, elimination of the aggregated bias does not always lead to the best performance. 
% To the best of our knowledge, that is the first case where it is proven on complex environments which highlights the important pitfalls of overestimation control.
\end{abstract}

\section{Introduction}
% TODO rewrite the beginning
An accurate estimation of a state-action value function~(Q-function) is essential to off-policy reinforcement learning.
One common challenge is overestimation bias, which introduces errors into value function estimates~\citep{thrun1993issues, hasselt2010double}. 
The bias emerges from imprecise maximization in temporal-difference learning, which leads to error accumulation~\citep{fujimoto2018addressing}.
As a consequence of the bias, sub-optimal actions may receive erroneous high-value estimates. 
This is believed to slow down the training and leads to poor results.
The desire for bias correction has inspired a number of high-performing reinforcement learning algorithms \citep{fujimoto2018addressing,lan2020maxmin, kuznetsov2020controlling}.
However, to adjust the magnitude of control, these algorithms usually require outer-loop hyperparameter tuning, effectively impairing the sample efficiency of reinforcement learning algorithms.

This paper proposes a novel approach inspired by Double Deep Q-Learning (DDQN) \cite{hasselt2016double} algorithm for discrete reinforcement learning. 
The authors proposed to select actions and assess them with two different Q-networks.
Each of the networks has different uncorrelated errors in the estimation of the true Q-function, and therefore the conditions for the overestimation bias's occurrence are not met.
Similarly, we can modify the algorithms that stem from Deep Deterministic Policy Gradient (DDPG) \cite{van2016deep} and are used in continuous RL: Truncated Quantile Critics (TQC) \cite{kuznetsov2020controlling}, Soft Actor Critic (SAC) \cite{haarnoja2018softapp}, Twin Delayed DDPG (TD3) \cite{fujimoto2018addressing}. 
All of them already have at least two Q-networks in them. 
If we model the policy in a form of a mixture of distributions, for each component we can ensure that the policy assessment step and the policy optimization step are performed using separate Q-networks.
As with DDQN, we expect it would completely remove the causes for the bias’s existence. 
And indeed, our experiments, albeit preliminary so far, show that with the application of our algorithm, the overestimation decreases and the performance improves, if compared with the cases of no bias elimination techniques at all.

We base our algorithm on TQC and use it as a baseline.
TQC still outperforms the proposed method both in terms of performance and bias elimination, but the results are comparable. 
On the other hand, TQC requires a grid search for one hyperparameter to get the best results, while our method removes this hyperparameter and allows single-run optimization on a new environment.

Overall our contributions are the following:
\begin{enumerate}
    \item We propose a method that deals with the problem of overestimation bias in continuous off-policy reinforcement learning that is applicable to a broad class of algorithms.
    \item We conducted preliminary experiments that showed that it may be a promising direction for further research, as the method shows a near-SOTA performance while removing the need for the environment-specific hyperparameter.
\end{enumerate}

\section{Background}
\subsection{Overestimation bias}

\citet{thrun1993issues} elucidate the overestimation as a consequence of Jensen's inequality: the maximum of the Q-function over actions is not greater than the expected maximum of noisy (approximate) Q-function.

\pagebreak
    Specifically, for any action-dependent random noise $U(a)$ such that $ \forall a ~~ \Ep{U}{U(a)} = 0$,
    \begin{equation}
        \begin{aligned}
            \max_a Q(s,a) 
            &=   
            \max_a \Ep{U}{Q(s, a) + U(a)} \\  
            &\leq  
            \Ep{U}{\max_a \{ Q(s,a) + U(a) \} }.
        \end{aligned}
    \end{equation}

In continuous off-policy RL instead of simple maximization $\max_a Q(s, a)$ we have two steps: policy optimization \ref{eq_1} and substitution of this policy into the Q-network in TD-target \ref{eq_2}, which ultimately lead to the same results.

\begin{equation}
    \label{eq_1}
    \ex_{a \sim \pi(s)} Q(s, a) \rightarrow \max_\pi
\end{equation}
\begin{equation}
    \label{eq_2}
    Q(s, a) \leftarrow \ex_{r, s' \sim p(s, a)} \ex_{a' \sim \pi(s')}\left[r + \gamma Q^{\text{target}}(s', a')\right]
\end{equation}
    
In practice, the noise $U(a)$ may arise for various reasons, such as spontaneous errors in function approximation, Q-function invalidation due to ongoing policy optimization, stochasticity of environment, etc. 
Off-policy algorithms grounded in temporal difference learning are especially sensitive to approximation errors since errors are propagated backward through episodic time and accumulate over the learning process.

\begin{table}
    % \begin{minipage}%${0.5\linewidth}
        \resizebox{0.5\textwidth}{!}{%
        \begin{tabular}{l l}
        \bf Notation & \bf Description \\
        \toprule
        ~~~~$\pi$ & Explicit policy of the agent \\
        \multirow{2}{*}{~~~~$\pi_{\phi_i}$} & $i$-th component of a mixture policy, \\
        & parameterized with the weights $\phi_i$ \\
        ~~~~$\mathcal{D}$ & Replay buffer \\
        \midrule
        \bf Q-functions\\
        ~~~~$Q^\pi$ & State-action value function for policy $\pi$\\
        \multirow{2}{*}{~~~~$Q_{\psi}$}  & Approximation to state-action value function,\\
        & Q-network with the weights $\psi$ \\
        ~~~~$Q_{\psi'}$ & Target Q-network\\
        ~~~~$N$ & Number of value function approximations\\
        ~~~~$Q_{\psi_i}$ & $i$-th approximation out of $N$\\
        % ~~~~$\hat{Q}_{t}$ & approximation at step $t$ of optimization\\
        % ~~~~$\tilde{Q}$ & Estimate of state-action value function\\
        \midrule
        \bf Distributional RL\\
        ~~~~$Z^\pi$ & Distributional state-action value function\\
        \multirow{2}{*}{~~~~$Z_{\psi_i}$} &  Approximation to distributional value function\\
        & returning $M$ scalars as the distribution quantiles\\
        ~~~~$x_{(i)}$ & $i$-th element of $x$ sorted in ascending order \\
        \bottomrule
        \end{tabular}
        }
        \caption{Notation used in the paper.}
        \label{tab:notation}
	% \end{minipage}
\end{table}

\subsection{Twin Delayed DDPG}
\label{prer:td3}
In \cite{fujimoto2018addressing} to combat overestimation bias the authors introduced Twin Delayed Deep Determenistic policy gradient (TD3). The main idea is to have two approximations of the Q-function and perform 'cliiping', where we take a minimum over them while constructing the TD-learning target.

\begin{equation}
y(r, s') = r + \gamma \min_{i=1,2} Q_{\psi'_n}(s', a'), \quad a' \sim \pi(s')
\end{equation}
\begin{equation}
\ex_{s,a,r,s' \sim \mathcal{D}}(Q_{\psi_i}(s, a) -y(r, s'))^2 \rightarrow \min_{\psi_n}
\end{equation}

This clipping removes all random spikes in the function landscape and heuristically fixes the problem to some extent.

TD3 also has a number of other modifications over the regular Deep Determenistic Policy Gradient (DDPG), but they are not important in this context.

% \subsection{Soft Actor Critic}
% \todo{Not sure this section is required}
\subsection{Truncated Quantile Critics}
\label{prer:tqc}
Another approach to fixing the problem of overestimation was taken by \cite{kuznetsov2020controlling}. 
Here the authors modify Soft Actor Critic \cite{haarnoja2018softapp} to have a distributional Q-function with explicit quantiles. 
Having multiple Q-functions of this type allows them to substitute clipping for truncation of the distribution that is used as TD-target.
Truncation here is zeroing out the probability of a number of the largest returns, or removing the right tail of the distribution. 
This process decreases the mean expected return.

Each $Z_{\psi_n}$ maps each $(s, a)$ to a probability distribution 
\vspace{-0.3cm}
\begin{equation}
    Z_{\psi_n}(s, a) := \frac{1}{M} \sum_{m=1}^{M} \delta \left(\theta_{\psi_n}^m(s, a) \right) ,
\end{equation}
supported on atoms $ \theta_{\psi_n}^1(s,a), \dots, \theta_{\psi_n}^M(s,a) $ .

They pool the atoms of the distributions $Z_{\psi_1}(s',a'), \dots, Z_{\psi_N}(s',a')$ into a set  
\begin{equation}
\mathcal{Z}(s',a') := \{\theta_{\psi_n}^m(s', a') \mid  n \in [1..N], m \in [1..M] \}
\end{equation}
and denote elements of $\mathcal{Z}(s',a')$ sorted in ascending order by $z_{(i)}(s',a')$, with $i \in [1..MN]$.

The $kN$ smallest elements of $\mathcal{Z}(s',a')$ define atoms 
\begin{equation}
\label{eq:tqc_atoms}
  y_i (s, a) := r(s,a) + \gamma z_{(i)}(s', a')
\end{equation}
of the target distribution
\vspace{-0.3cm}
\begin{equation}
Y(s,a) := \frac{1}{kN} \sum_{i=1}^{kN} \delta \left( y_i (s, a) \right) .
\end{equation}
\vspace{-0.3cm}

They approximate the $\tau_m, m \in [1..M]$ quantiles of $Y(s,a)$ with  $ \theta_{\psi_n}^1(s,a), \dots, \theta_{\psi_n}^M(s,a) $  by minimizing the loss
\begin{equation}\label{eq:value_loss}
    J_Z (\psi_n) = 
    \Ep{\mathcal{D}, \pi} { 
             \mathcal{L}^k(s_t,a_t; \psi_n)
    }, 
\end{equation}
over the parameters $\psi_n$, where 
\begin{equation}
    \mathcal{L}^k(s,a; \psi_n) = \frac{1}{kNM} \sum_{m=1}^M \sum_{i=1}^{kN} \rho_{ \tau_m}^H(y_i(s, a) - \theta_{\psi_n}^m(s,a)).
\end{equation}

\begin{equation}
    \mathcal{L}^k(s,a; \psi_n) \rightarrow \min_{\psi_n}
\end{equation}

As TQC is based on SAC, it also has entropy terms in policy optimization and TD-learning, but they are unrelated to the main topic of the paper, so we omitted them from the formulae.

\subsection{Double Q-Learning}
In discrete RL the main algorithm to fix the overestimation bias is Double Q-Learning \cite{hasselt2016double}. 

Here instead of constructing TD-target naively

\begin{equation}
    y = r + \gamma \max_{a'} Q_\psi (s', a')
\end{equation}

the authors use the main Q-function to select the action and then the target Q-function to properly assess it

\begin{equation}
    y = r + \gamma Q_{\psi'} (s', \arg\max_{a'}Q_\psi(s', a'))
\end{equation}

Therefore the action selection and the action assessment are uncorrelated, and the overestimation bias is basically nonexistent in such environments as Atari.

\section{Continuous Double Q-learning}
\label{sec:cdq}
In our work, we aim to eliminate heuristics in overestimation bias correction in continuous RL (\ref{prer:td3}, \ref{prer:tqc}). Their main problem is that they have hyperparameters (number of truncated quantiles and number of networks to take minimum over) that are not clear how to tune and what the optimal value should be. Instead, we are trying to reach the same properties the DDQN has: high effectiveness and a theoretical (or at least a common sense) justification. 

In this section we introduce Continuous Double Q-Learning algorithm (CDQ), that is designed to achieve these goals.
We describe CDQ as a modification of a general DDPG-like algorithm, while in reality we modify TQC and conduct the experiments with that algorithm.

Main idea of the algorithm is to have the policy $\pi$ in a form of a mixture of two components. Then as with DDQN each component would be assessed and optimized by two different Q-networks.
\begin{equation}
    \pi(a) = \dfrac{1}{2}(\pi_{\phi_1}(a) + \pi_{\phi_2}(a))
\end{equation}

We have two Q networks in total, each of them is an approximation of the real $Q^\pi$:
\begin{equation}
    Q_{\psi_1} \approx Q^{\pi}, \quad Q_{\psi_2} \approx Q^{\pi}
\end{equation}

Policy optimization is the same as in DDPG and SAC, we differentiate through the Q-function:

\begin{equation}
    \ex_s \ex_{a \sim \pi_\phi(s)} Q_\psi(s, a) \rightarrow \max_\phi 
\end{equation}

But each component of the policy is optimized with its own Q-network:

% \begin{equation}
%     \ex_s \dfrac{1}{2} \sum_{i=1}^{2}\ex_{a \sim \pi_{\phi_i}(s)} Q^\pi(s, a) \rightarrow \max_{\phi_i} 
% \end{equation}

\begin{equation}
    \ex_s \dfrac{1}{2} \sum_{i=1}^{2}\ex_{a \sim \pi_{\phi_i}(s)} Q_{\psi_i}(s, a) \rightarrow \max_{\phi_i}
    \label{for:policy_max}
\end{equation}

As for the action assessment in TD-target, we also modify the classic formula to include two Q-networks, but now we swap them, and each policy component is assessed by the Q-network that is different from the one used in policy optimization \ref{for:policy_max}.

\begin{equation}
    y(r,s') = r + \gamma Q^\pi(s', \pi(s'))
\end{equation}

\begin{equation}
    y(r, s') = r + \dfrac{1}{2} \gamma \sum\limits_{i=1}^{2}  Q_{\psi_i}(s', \pi_{\phi_{2-i}}(s'))
\end{equation}

\begin{equation}
    \ex_{s, a, r, s' \sim \mathcal{D}}(Q_{\psi_i}(s, a) - y(r, s'))^2 \rightarrow \min_{\psi_i} \quad i = 1, 2
\end{equation}

Making such changes in TD-learning and policy optimization ensures that for each policy component $\pi_i$ the Q-networks that used for action assessment and policy optimization are different. Theoretically, it removes the ground for overestimation bias completely.

It is important to note that \cite{fujimoto2018addressing} already tried applying DDQN to continuous RL, and they came to similar formulae with two policies, two Q-networks, and cross-optimization. However, the description is very brief, the code is not released, and it is unclear how these two policies form the final policy and which policy performs explorative actions. The authors came to a conclusion that the performance of this version of DDQN is worse than that of TD3, the algorithm they proposed. 
We think that it is possible to make a version of Double Q-Learning for continuous RL, but even the failure of doing so may shed a light on the key differences between overestimation bias mechanisms in continuous and discrete RL's.

\begin{figure*}[!t]
\centering
\includegraphics[width=\textwidth]{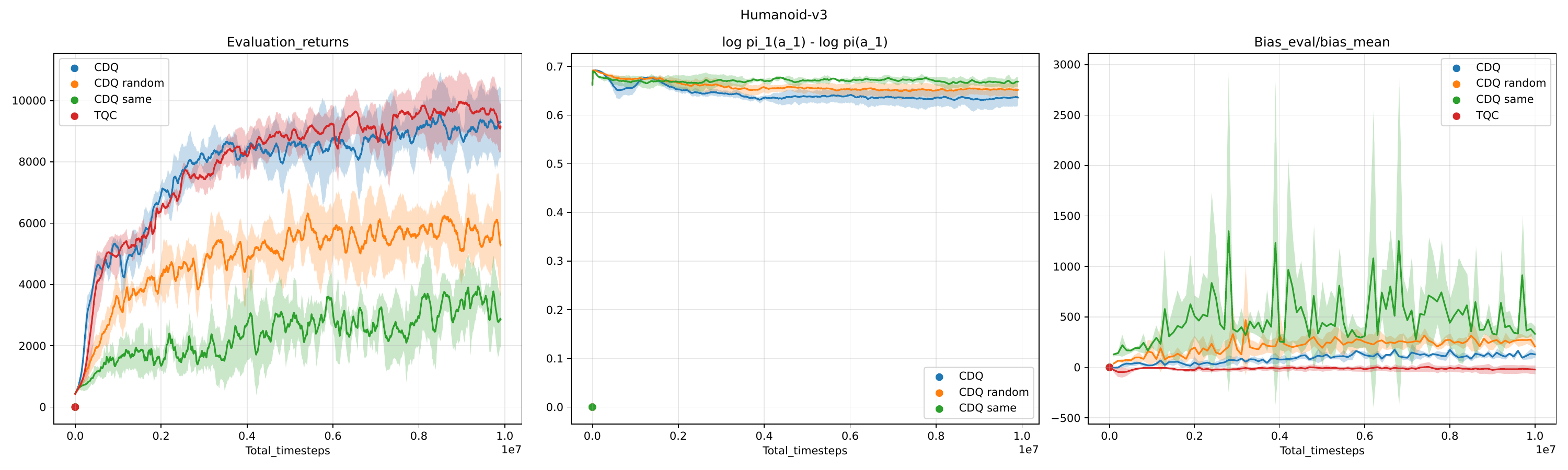}
\includegraphics[width=\textwidth]{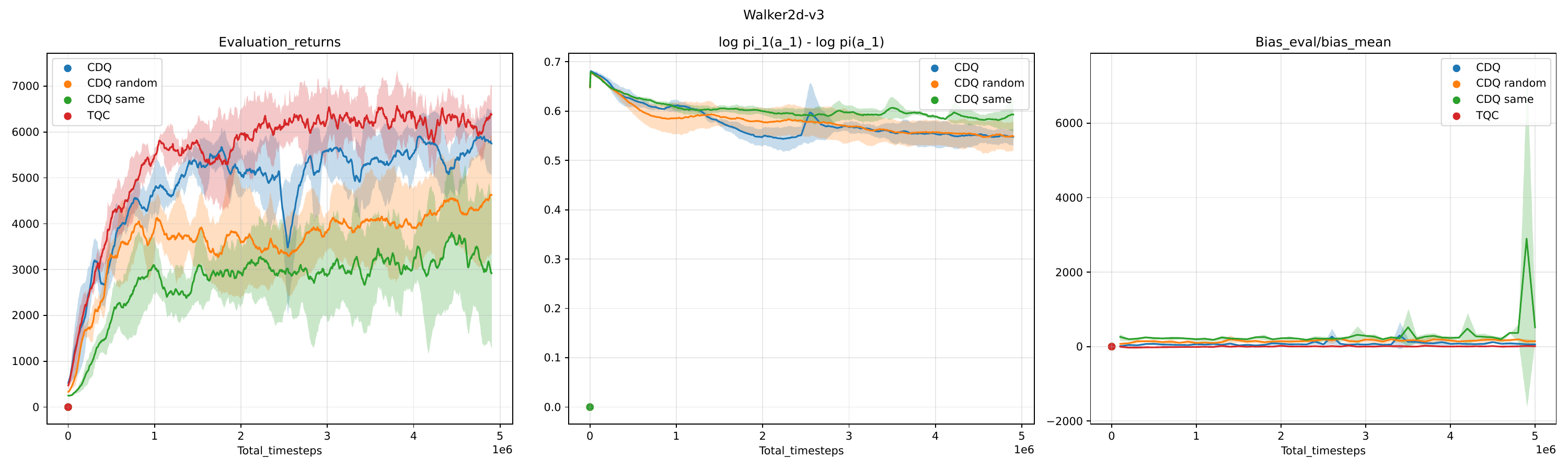}
\caption{
Comparison of TQC, CDQ and ablation variants CDQ-random, CDQ-same. The first row is MuJoCo 'Humanoid-v3' environment, the second row is 'Walker2d-v3'. The first column shows evaluation returns, the second column shows difference $\log \pi_0(a_0) - \log \pi(a_0)$, the third column -- on-policy overestimation bias.
}
\label{fig:exps}
\end{figure*}

\section{Experiments}
We conducted only preliminary experiments on two MuJoCo environments: Humanoid and Walker2d. In these experiments we compared baseline TQC with its modified version, CDQ. For ablation purposes we also have two versions of CDQ, that gradually remove the overestimation bias alleviation technique.

All algorithms have two policy components $\pi_{\phi_i}$ and two Q-functions $Q_{\psi_i}$.
\begin{enumerate}
    \item CDQ -- this algorithm is a modification of TQC (\ref{prer:tqc}) with the modifications described in section \ref{sec:cdq}. Each $i$-th policy component is optimized with the corresponding Q-function $Q_{\psi_i}$, and evaluated with the opposite Q-function $Q_{\psi_{2-i}}$, $i=1,2$.
    \item CDQ-random -- in this version each $\pi_i$ is optimized with $Q_{\psi_i}$, but evaluated randomly with $Q_{\psi_1}$ and $Q_{\psi_2}$ with equal probability.
    \item CDQ-same: each policy component $\pi_i$ is both optimized and evaluated by the same corresponding Q-function $Q_{\psi_i}$.
\end{enumerate}

The expectation is that CDQ should be free of overestimation bias, CDQ-same should fully exhibit the phenomenon of overestimation bias, and CDQ-random should be in the middle.

The following conclusions could be drawn from the limited experimental results in figure \ref{fig:exps}:

\begin{enumerate}
    \item CDQ shows almost the same performance in terms of evaluation returns, as TQC. While TQC achieves this performance after grid search on its parameter, CDQ requires only one run.
    \item CDQ indeed decreases overestimation bias, but it still present. TQC with the optimal hyperparameters shows better result at bias elimination. This is a contradiction to our expectation. It is probably that additional modification are required in order to achieve zero average bias, such as keeping two separate replay buffers for the optimization of different policies and Q-networks.
    \item While in theory $\pi_{\phi_1}$ and $\pi_{\phi_2}$ should exhibit similar behavior, we can see that there is significant difference between $\log \pi_1(a_1)$ and $\log \pi(a_1)$, $a_1 \sim \pi_{\phi_1}$. The difference starts at $\log 2 \approx 0.69$ and slowly decreases, and given that
    \begin{equation}
    \ex_{a_1 \sim \pi_1} \left[\log \pi_1(a_1) - \log \pi (a_1)\right] =
    \end{equation}
    $$
    \ex_{a_1 \sim \pi_1} \log \dfrac{\pi_1(a_1)}{\pi_1(a_1) + \pi_2(a_1)} + \log 2 = 
    $$
    \begin{equation}
    \log 2 - \ex_{a_1 \sim \pi_1} \log \left(1 + \dfrac{\pi_2(a_1)}{\pi_1(a_1)}\right) \approx \log 2 \Rightarrow
    \end{equation}
    \begin{equation}
    \pi_2(a_1) \approx 0, \quad a_1 \sim \pi_1
    \end{equation}
    we can see that in the beginning the policy components are quite different, and almost don't converge during the learning.
\end{enumerate}

\section{Conclusions}
In this short paper we propose a way to apply Double Q-Learning idea to Continuous Reinforcement Learning. 
% It is a novel idea, that wasn't researched in the literature. 
This idea is not researched well in the literature, while it could improve the overall performance of state of the art RL methods and get rid of the need in heuristic approaches such as clipping \cite{fujimoto2018addressing} and truncation \cite{kuznetsov2020controlling}.

We expected this approach to solve the overestimation bias problem completely, but the experiments show that while the situation has improved significantly over the naive algorithm, some degree of overestimation still remains. 
The method's overall performance is also not as good as the baseline with the heuristic mechanism and the optimal hyperparameters. 
At the same time, the experiments show a conspicuous correlation between the overestimation bias and the performance. 

% We see the following directions for developing the proposed idea:
As the following direction for research, we think it is important to understand why the bias is not negated completely. 
\begin{enumerate}
    \item Confirm rigorously the theoretical correctness of the algorithm.
    \item Conduct experiments with a separate replay buffer for each policy.
\end{enumerate}

\bibliography{main}
\bibliographystyle{icml2021}


\end{document}